\documentclass[lettersize,journal]{IEEEtran}
\usepackage{amsthm}

\usepackage{amsmath,amsfonts,mathrsfs}
\usepackage{amsmath}               
\usepackage{svg}
\usepackage{algorithm} 
\usepackage{algpseudocode} 
\usepackage{array}
\usepackage[caption=false,font=normalsize,labelfont=sf,textfont=sf]{subfig}
\usepackage{textcomp}
\usepackage{stfloats}
\usepackage{url}
\usepackage{verbatim}
\usepackage{graphicx}
\usepackage{cite}
\usepackage[utf8]{inputenc}
\usepackage{graphicx}
\usepackage{epstopdf}
\usepackage{etoolbox}
\usepackage{cite}
\usepackage{picinpar}
\usepackage{flushend}
\usepackage{bm}
\usepackage{xcolor}
\usepackage{indentfirst}
\usepackage{balance}
\usepackage[autostyle]{csquotes}
\expandafter\def\expandafter\normalsize\expandafter{%
    \normalsize%
    \setlength\abovedisplayskip{4pt}%
    \setlength\belowdisplayskip{4pt}%
    \setlength\abovedisplayshortskip{0pt}%
    \setlength\belowdisplayshortskip{0pt}%
}

\begin{document}

\title{Fractional-order Modeling for Nonlinear Soft Actuators via Particle Swarm Optimization}

\author{Wu-Te Yang, Masayoshi Tomizuka,~\IEEEmembership{IEEE Life Fellow}
\thanks{The author are with the Department of Mechanical Engineering,
        University of California, Berkeley, MSC Lab, USA
        {\tt\small wtyang; tomizuka@berkeley.edu}}        
        }



\maketitle

\begin{abstract}
Modeling soft pneumatic actuators with high precision remains a fundamental challenge due to their highly nonlinear and compliant characteristics. This paper proposes an innovative modeling framework based on fractional-order differential equations (FODEs) to accurately capture the dynamic behavior of soft materials. The unknown parameters within the fractional-order model are identified using particle swarm optimization (PSO), enabling parameter estimation directly from experimental data without reliance on pre-established material databases or empirical constitutive laws. The proposed approach effectively represents the complex deformation phenomena inherent in soft actuators. Experimental results validate the accuracy and robustness of the developed model, demonstrating improvement in predictive performance compared to conventional modeling techniques. The presented framework provides a data-efficient and database-independent solution for soft actuator modeling, advancing the precision and adaptability of soft robotic system design.
\end{abstract}


\section{Introduction}
\IEEEPARstart{S}{oft} robots have attracted attention in recent years due to their superior degrees of freedom and compliance compared to traditional rigid robots. Their unique capabilities enable them to operate in challenging and unstructured environments including extreme environments~\cite{Tang2023unknown}, to deliver delicate components in medical applications~\cite{alici2018bending}, and to handle fruits in the food industry~\cite{dai2023soft}. The motion of soft robots primarily relies on soft actuators because of ease of fabrication, cost-effectiveness, and high power density~\cite{Tolley2018design, c31}. Despite these advantages, soft actuators, due to their nonlinear structures, pose challenges for dynamic modeling~\cite{yang2024model}

Soft-robot dynamic modeling has been widely explored using PCC~\cite{c14}, Cosserat rod~\cite{xun2024cosserat}, PDE-based~\cite{zheng2022pde}, and Lagrangian methods~\cite{c16}. Although accurate, these approaches often involve complex structures or labor-intensive modeling. To improve tractability, several works~\cite{c10, c11} approximate soft pneumatic actuators as second-order systems identified by damping ratios and natural frequencies. However, linear models lose accuracy at large bending angles, and nonlinear second-order models degrade even at smaller deformations~\cite{yang2024model}. More recently, fractional-order differential equations have been introduced to capture the memory-like behavior of soft robots~\cite{Di2011fo, morena2024fo}.

Fractional-order differential equation (FODE) models offer an effective balance between complexity and accuracy in soft-robot modeling. Physically, properties such as viscoelasticity and material memory describe behaviors that lie between those of a purely elastic solid and a purely fluid medium~\cite{hert2020mechanics}. Mathematically, these behaviors are often represented using classical exponential equations; however, soft materials may respond at rates that are either faster or slower than those predicted by exponential laws. This mismatch motivates the use of fractional models, which naturally capture such intermediate or history-dependent dynamics. By allowing non-integer derivative orders, fractional models flexibly bridge the gap between purely elastic and purely viscous responses~\cite{morena2024fo}, providing higher accuracy with fewer parameters compared to high-order integer models.

This study models the dynamics of soft pneumatic actuators (SPAs) using fractional-order differential equations (FODEs), with fractional orders and system parameters identified via particle swarm optimization (PSO). The model is validated on actuators made from different materials and compared with recent methods. The main contributions are: (1) introducing a FODE-based modeling framework for SPAs with PSO-based parameter identification; (2) removing the need for tensile testing and material databases; and (3) experimentally validating the method and analyzing PSO limitations, including the minimum data required for accurate modeling

To position our contributions, this research is compared with recent works. Yang et al.\cite{yang2024model} proposed a nonlinear second-order dynamic equation, with system parameters identified through a data-driven approach and materials' database, has been applied to model SPAs. However, the present study aims to eliminate the reliance on such databases by employing an AI-based fractional-order model to describe the behavior of SPAs. Morena et al.~\cite{morena2024fo} proposed a fractional-order modeling approach for hydrogel-based soft pneumatic bending actuators, in which the system parameters were determined through empirical identification. In contrast, our work employs an optimization algorithm to identify the system parameters, offering improved efficiency and accuracy. Overall, this work presents a fractional-order modeling framework for SPAs, achieving improved accuracy and efficiency, and providing insight into the limitations of PSO and physical meaning of fractional dynamics.

The remainder of this paper is organized as follows. Section~\ref{sec2} introduces the fractional-order model. Section~\ref{sec3} details the PSO algorithm. Section~\ref{sec4} evaluates the effectiveness of the fractional-order model. Lastly, section~\ref{sec5} discusses the results and concludes the work.

\section{Fractional-Order Dynamic Model}\label{sec2}
\subsection{Soft Pneumatic Actuator}\label{subsec21}
\begin{figure}[t]
    \centering
    \includegraphics[width=190pt]{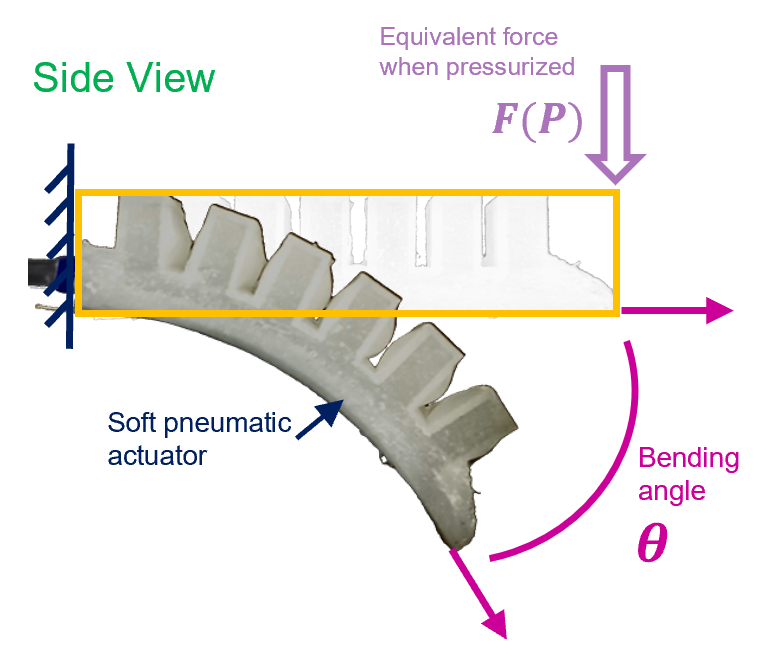}
    \caption{The soft pneumatic actuator is undergoing bending $\theta$ with equivalent force $F(P)$ when it is pressurized by an air pump.}
    \label{fig: 1}
    \vspace{-0.15in}
\end{figure}

The soft pneumatic actuator used in this study follows the optimized design proposed by Yang et al.~\cite{yang2024optimized}. Previous work has demonstrated that employing nonlinear second-order models~\cite{yang2024model} can improve the modeling accuracy of this actuators. SPAs have several discrete chambers, and their nonlinear structure is approximated as a cantilever beam. The second-order model is obtained on the basis of structure approximation as Fig.~\ref{fig: 1}.

Previously, the bending dynamic equation of the soft pneumatic actuator is derived and the nonlinear dynamic model of the soft actuator is given by~\cite{yang2024model}:

\begin{align}
    \begin{split}
    {M_{eq}} \ddot{\theta} + C_{n_p}{\dot \theta} + K_{n_p} {\theta}^{{n_p}+\Delta {n_p}} = F
    \label{eqn: 5}
    \end{split}
\end{align}
\begin{align}
    \begin{split}
    K_{n_p} = (\frac{{n_p}+1}{n_p})^{n_p}(\frac{EI_{n_p}}{L_0^{n_p+1}})
    \label{eqn: 2}
    \end{split}
\end{align}
where $n_p$ is a number which is usually greater than $1$ for soft materials, $\Delta n_p$ represents the perturbation of soft materials, $M_{eq}$ is the equivalent mass of the soft actuator, $C_{n_p}$ is the damper of the soft actuator, $K_{n_p}$ is the spring constant obtained from (\ref{eqn: 2}). 

Moreover, $E$ represents the Young's modulus, $\theta$ is the bending angle as Fig.~\ref{fig: 1}, $L_0$ is the initial length of the structure, $I_{n_p}$ is the modified moment of inertia for a large deflection component, and it is expressed as

\begin{align}
    \begin{split}
    {I_{n_p}} = (\frac{1}{2})^{n_p}(\frac{1}{2+n_p})b{h}^{(2+n_p)} 
    \label{eqn: 3}
    \end{split}
\end{align}

The nonlinear second-order model achieves high accuracy in various applications. However, determining the fractional order $n_p$ can be challenging for certain soft materials due to the lack of comprehensive databases containing material properties such as Young's modulus, tensile strength, and viscoelastic parameters. Furthermore, the nonlinear second-order model may exhibit deviations at small bending angles. Given the limitations of this method, it is necessary to propose a different approach. Some experimental results are discussed in Sec.~\ref{sec43}.

\subsection{Fractional Differential Equation}\label{subsec22}
Physical systems are typically modeled using integer-order differential equations. However, when systems exhibit memory, creep, or hysteresis effects such as viscoelastic behaviors, integer-order models may fail to capture their dynamics accurately~\cite{kempfle2001fo,Di2011fo}. These properties and behaviors lie between those of a purely elastic solid and those of a pure fluid medium. In such cases, fractional-order differential equations, introduced by Riemann and Liouville~\cite{podlubny1998fde}, are often employed to more effectively represent these phenomena.

\subsubsection{Definition of Fractional Differential Equations}\label{subsec221}
The Riemann-Liouville (RL) definition~\cite{chen2009fo} is commonly used for general fractional differential equations. The definition of the fractional operator is:

\begin{align}
    \begin{split}
    {{_a}D_{t}^{r}f(t)} = \frac{1}{\Gamma(n-r)} \frac{d^n}{dt^n} \int_{a}^{t} \frac{f(\tau)}{(t-\tau)^{r-n+1} } d\tau
    \label{eqn: 6}
    \end{split}
\end{align}
where $r$ is the fractional-order and $r \in (n-1, n]$, $\Gamma (\cdot)$ is the Gamma function, and $a$ and $t$ are the lower and upper limits of the integration.

Since the Laplace transform is widely used in the analysis of system differential equations for dynamic simulations and controller design, the Laplace transform of the RL fractional differential equation~(\ref{eqn: 6}) is defined as~\cite{chen2009fo}: 

\begin{align}
    \begin{split}
    \int_{0}^{\infty} e^{-st}{_0}D_{t}^r f(t)dt = s^r F(s) - \sum_{k=0}^{n-1} s^k{_0}D_{t}^{r-k-1}f(t) \Big|_{t=0}
    \label{eqn: 7}
    \end{split}
\end{align}
where $s=j\omega$ and $r \in (n-1, n]$.

When $r = n$ and $n$ is an integer, the (\ref{eqn: 6}) becomes the integer-order differential equation. When $r=0$, the fractional differential operator (\ref{eqn: 6}) is the identity operator ${_0}D_{t}^{0}f(t)=f(t)$. Both fractional differentiation and fractional integration are linear operations~\cite{chen2009fo}.

\subsubsection{Fractional-order Dynamic Systems}\label{subsec222}
With the definitions of the fractional operator~(\ref{eqn: 6}), the behavior of a fractional-order dynamic system can be represented by the following fractional differential equation~\cite{podlubny1998fde}:

\begin{align}
    \begin{split}
    {a_n}&D^{\alpha_{n}}y(t) + {a_{n-1}}D^{\alpha_{n-1}}y(t)+\dots+{a_0}D^{\alpha_{0}}y(t) \\ =& {b_m}D^{\beta_{m}}u(t) + {b_{m-1}}D^{\beta{m-1}}u(t) + \dots +{b_0}D^{\beta_{0}}u(t)
    \label{eqn: 9}
    \end{split}
\end{align}
where $D^{\gamma}\equiv{_0}D_{t}^{\gamma}$, ${a_n} \dots {a_0}$ and ${b_m} \dots {b_0}$ are constants, and ${\alpha_n} \dots {\alpha_0}$ and ${\beta_n} \dots {\beta_0} \in \mathbb{R}_{+}$. Without loss of generality, it is assumed that $\alpha_n > \alpha_{n-1} > \dots>\alpha_0=0$ and $\beta_m>\beta_{m-1}>\dots>\beta_0=0$.

We take Laplace transformation of Eq.~(\ref{eqn: 9}) by using (\ref{eqn: 7}) and set initial conditions as 0. The fractional-order transfer function $G(s)$ is obtained and is described as:

\begin{align}
    \begin{split}
    G(s) = \frac{{b_m s^{\beta_m}}+\dots+{b_0}}{{a_n}s^{\alpha_n}+\dots+{a_0}}
    \label{eqn: 10}
    \end{split}
\end{align}

Equation (\ref{eqn: 10}) is the general form and this study considers a transfer function of order $\alpha_2$. Also, the input of SPA is assumed to be a constant source (air pressure), so the numerator is a constant~\cite{yang2024model}. The fractional-order transfer function is expressed as:

\begin{align}
    \begin{split}
    G(s) &= \frac{b_0}{{a_2}s^{\alpha_2}+{{a_1}s^{\alpha_1}}+{a_0}} \\
         &= \frac{b_0'}{s^{\alpha_2}+a_1's^{\alpha_1}+a_0'}
    \label{eqn: 11}
    \end{split}
\end{align}
where $0<\alpha_1<\alpha_2<2$. Since the coefficients $a_1'=a_1/a_2=2\zeta\omega_n, a_0'=a_0/a_2=\omega_n^2,$ and $b_0'=b_0/a_2$ are unknown, they are determined using the optimization method described in Sec.~\ref{sec3}. Then, Eq. (\ref{eqn: 11}) is used to describe the dynamics of soft pneumatic actuators in Sec.~\ref{sec43}.

\subsubsection{Analytical Solution of Fractional-order Equations}\label{subsec223}
The fractional-order dynamic equation such as (\ref{eqn: 11}) is defined previously. The analytical solution of it will be discussed to further analyze the systems' responses. The unit-step response is:

\begin{align}
    \begin{split}
    Y(s)= \frac{b_0'}{s^{\alpha_2}+a_1's^{\alpha_1}+a_0'} U(s)
    \label{eqn: 12}
    \end{split}
\end{align}
where $U(s)=\frac{1}{s}$. Thus,

\begin{align}
    \begin{split}
    Y(s)= \frac{b_0'}{s(s^{\alpha_2}+a_1's^{\alpha_1}+a_0')}
    \label{eqn: 13}
    \end{split}
\end{align}
The inverse Laplace transformation of (\ref{eqn: 13}) yields

\begin{align}
    \begin{split}
    y(t)= \mathcal{L}^{-1}\biggl\{\frac{b_0'}{s(s^{\alpha_2}+a_1's^{\alpha_1}+a_0')}\biggl\}
    \label{eqn: 14}
    \end{split}
\end{align}
The exact analytical form of (\ref{eqn: 14}) is given by~\cite{podlubny1998fde}

\begin{align}
    \begin{split}
    y(t)= {b_0}'\sum_{k=0}^{\infty}(-a_1')^k t^{(\alpha_2-\alpha_1)k+1}\it{E}_{\alpha_2,(\alpha_2-\alpha_1)k+2}^{k+1}(-a_0't^{\alpha_2})
    \label{eqn: 15}
    \end{split}
\end{align}
where $\it{E}_{\alpha,\beta}^{\gamma}(z)$ is the three-parameter Mittag-Leffler (ML) function and based on (\ref{eqn: 15}), $\alpha = \alpha_2$, $\beta=(\alpha_2-\alpha_1)k+2$, and $\gamma=k+1$. The ML function is defined as~\cite{haubold2011ml}:

\begin{align}
    \begin{split}
    \it{E}_{\alpha,\beta}^{\gamma}(z) = \sum_{n=0}^{\infty}\frac{(\gamma)_n z^n}{n!\Gamma(\alpha n+\beta)}
    \label{eqn: 16}
    \end{split}
\end{align}
where $(\gamma)_n$ is the Pochhammer symbol and $(\gamma)_n=\gamma(\gamma+1)(\gamma+2)\cdots(\gamma+n-1)$. Also, when $n=0$, $(\gamma)_0=1$. The unknown parameters in Eq.~(\ref{eqn: 11}) are searched by PSO discussed in Sec.~\ref{sec3}.

\section{Particle Swarm Optimization}\label{sec3}
Particle Swarm Optimization (PSO) is selected to search and determine unknown parameters in (\ref{eqn: 11}). PSO is a population-based stochastic optimization technique that offers some advantages~\cite{bratton2007pso}. First, PSO is conceptually simple and easy to implement, requiring only a few control parameters and no gradient information of the objective function. This makes it suitable for optimizing nonlinear, discontinuous, and non-differentiable problems like fractional-order equations. Second, due to its collective learning mechanism, combining individual experience and social sharing, PSO demonstrates strong global search capability and fast convergence in the early stages of optimization. Thus, it is selected for determining parameters in fractional-order equations.

In PSO, each particle represents a potential solution characterized by a position and a velocity in the search space. During each iteration, particles adjust their trajectories based on two key pieces of information: (1) each particle's best position found at the moment $(p_{pb})$ and (2) the global best position found by any particle in the swarm $(p_gb)$. The movement of each particle is governed by the following update equations~\cite{bratton2007pso}:

\begin{align}
    \begin{split}
    v_{i}^{t+1} = wv_{i}^{t} + c_{1}r_{1}(p_{pb}-x_{i}^{t})+c_{2}r_{2}(p_{gb}-x_{i}^t)
    \label{eqn: 17}
    \end{split}
\end{align}
\begin{align}
    \begin{split}
    x_{i}^{t+1} = x_{i}^{t}+v_{i}^{t+1}
    \label{eqn: 18}
    \end{split}
\end{align}
where $x_{i}$ and $v_{i}$ denote the position and velocity of the $i^{th}$ particle, $x_{i}$ contains the unknown parameters including $b_0',a_0',a_1', \alpha_1,$ and $\alpha_2$ in (\ref{eqn: 11}), $w$ is the inertia weight constant which determines how much should the particle keep on with its previous velocity,  $c_1$ and $c_2$ are cognitive and social learning coefficients, and $r_1$ and $r_2$ are random numbers uniformly distributed in $[0,1]$. As searching in the whole space may be inefficient, upper bound and lower bound are set for $x_{i}$. The upper and lower bounds in this study are shown in Sec.~\ref{sec41}.

The fitness function is defined as the root-mean-square error between real responses of SPAs, $y_{exp}(t)$, and $y(t)$ in (\ref{eqn: 15}):

\begin{align}
    \begin{split}
    f(x_i^{t+1}) = RMSE(y_{exp}(t) , y(t, x_i^{t+1}))
    \label{eqn: 19}
    \end{split}
\end{align}
The PSO pseudocode is shown below:
\begin{algorithm}
\caption{PSO Update Algorithm}\label{alg:pso}
    Randomly initialize Swarm population of $N$ particle $i$ within upper and lower bounds\\
    Select hyperparameter values $w$, $c_1$ and $c_2$
	\begin{algorithmic}[1]
		\For {each time step $t$}
			\For {each particle $i$ in the swarm}
				\State update $v_{i}^{t+1}$ and $x_{i}^{t+1}$ using Eq.~(\ref{eqn: 17}) \&(\ref{eqn: 18})
				\State Compute particle fitness $f(x_i^{t+1})$ using Eq.~(\ref{eqn: 19})
                \State Update $p_{pb}$ and $p_{gb}$
			\EndFor
		\EndFor
        ~Return best particle of Swarm
	\end{algorithmic} 
\end{algorithm}

The results of the parameter search for the soft actuators are demonstrated in Sec.~\ref{sec41}. The limitations of the PSO, including the minimum data required, are also discussed in Sec.~\ref{sec41}. 

\section{Experimental Evaluation}\label{sec4}

\subsection{Fractional-order Models Setup}\label{sec41}
Two soft actuators which are made of different soft materials, Ecoflex\textregistered Dragon Skin 20 and Ecoflex\textregistered Dragon Skin FX-Pro respectively, are used to validate the fractional-order differential equations discussed in Sec.~\ref{subsec22}. The fractional-order models will be compared with nonlinear model (\ref{eqn: 5}) which is proposed recently~\cite{yang2024model}. 

The parameters of fractional-order model~(\ref{eqn: 11}) for two soft materials are searched by PSO and are shown in Table~\ref{tab:Table I}. The swarm particle size is set as 200 and the iteration is 10. The constraints of each parameter are as follow: $\alpha_2$ ranges from 1.1 to 1.9, $\alpha_1$ from 1.0 to 1.5, $a_{0}'= \omega_n^2$ from 1 to 2, $a_{1}' = 2{\zeta}{\omega_n}$ from 0.5 to 1, and $b_0'$ from  1 to 2. The upper and lower bounds of $\alpha_2$ and $\alpha_1$ are determined by referencing~\cite{hert2020mechanics}. Similarly, the bounds for the damping ratio, natural frequency, and $b_0'$ are selected with reference to Eq.~(\ref{eqn: 5}) in~\cite{yang2024model}, which can be restated as: 
\begin{align}
    \begin{split}
    \ddot{\theta} + 2\zeta\omega_n{\dot \theta} + \omega_n^2 {\theta}^{{n_p}+\Delta {n_p}} = {F}/{M_{eq}}
    \label{eqn: 22}
    \end{split}
\end{align}

Due to the inherent uncertainty of soft materials, the parameters listed in Table~\ref{tab:Table I} are identified using PSO based on multiple experimental datasets. To ensure reliable parameter estimation, at least 4 datasets (each containing 50 data points) are used. Detail discussion can be seen in Sec.~\ref{subsection513}.

\begin{table}[http] 
\centering
\caption{\label{tab:Table I}The parameters in (\ref{eqn: 15}) are searched by PSO.}
\begin{tabular}{|c | c | c| c| c| c|}
\hline
\textbf{Material Name} & $\alpha_2$ & $\alpha_1$ & $a_1'$ & $a_0'$ & $b_0'$\\ [0.5ex]
\hline
Dragon Skin 20 & 1.406 & 1.196 & 0.794 & 1.638 & 0.934  \\ [0.5ex]
\hline
Dragon Skin FX-Pro & 1.424 & 1.170 & 0.986 & 2.103  & 0.728\\ [0.5ex] 
\hline
\end{tabular}
\vspace{-0.2in}
\end{table}

\subsection{Experimental Setup}\label{sec42}
\begin{figure}[t]
    \centering
    \includegraphics[width=195pt]{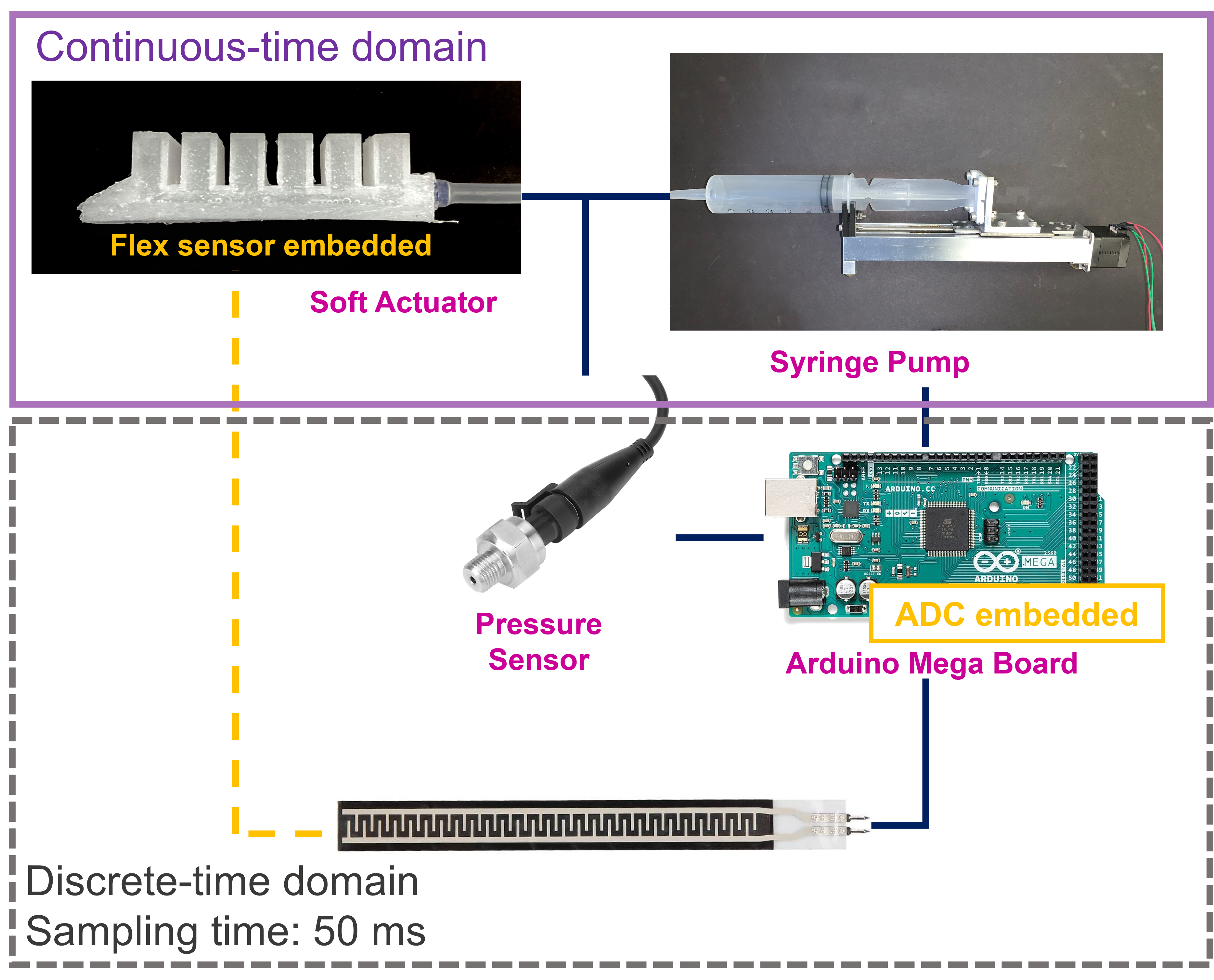}
    \caption{The schematic of experimental setup.}
    \label{fig: 3}
    \vspace{-0.15in}
\end{figure}

Figure~\ref{fig: 3} illustrates the control block diagram and experimental setup~\cite{yang2024model}. The soft actuators are driven by a custom-designed syringe pump. For open-loop control, an air pressure sensor (Walfront, Lewes, DE) with a range of 0–80 psi is used to monitor the actuator pressure. A flex sensor (Walfront, Lewes, DE) integrated within the actuator measures the bending angle, enabling feedback control. Both sensors are interfaced with an Arduino MEGA 2560 microcontroller (SparkFun Electronics, Niwot, CO), based on the Microchip ATmega 2560 platform, which in turn is connected to a computer for data acquisition.

\subsection{System Response Test}\label{sec43}
Step response tests are conducted to validate the effectiveness of the fractional-order model discussed in Sec.~\ref{subsec22} Using the parameters listed in Table~\ref{tab:Table I}, the fractional-order model output for the soft actuators is visualized and compared with the experimental data.
The bending angle of each soft actuator is defined as illustrated in Fig.~\ref{fig: 1}. Each actuator is embedded with a flex sensor, which is a resistive-type sensor. The variations in resistance are mapped to corresponding bending angles. Since the actuators are fabricated from different soft materials, the one made of Dragon Skin~20 is referred to as Design~1, whereas the one made of Dragon Skin~FX-Pro is referred to as Design~2.

\subsubsection{Design 1 Test} 
\begin{figure}[t]
    \centering
    \includegraphics[width=205pt]{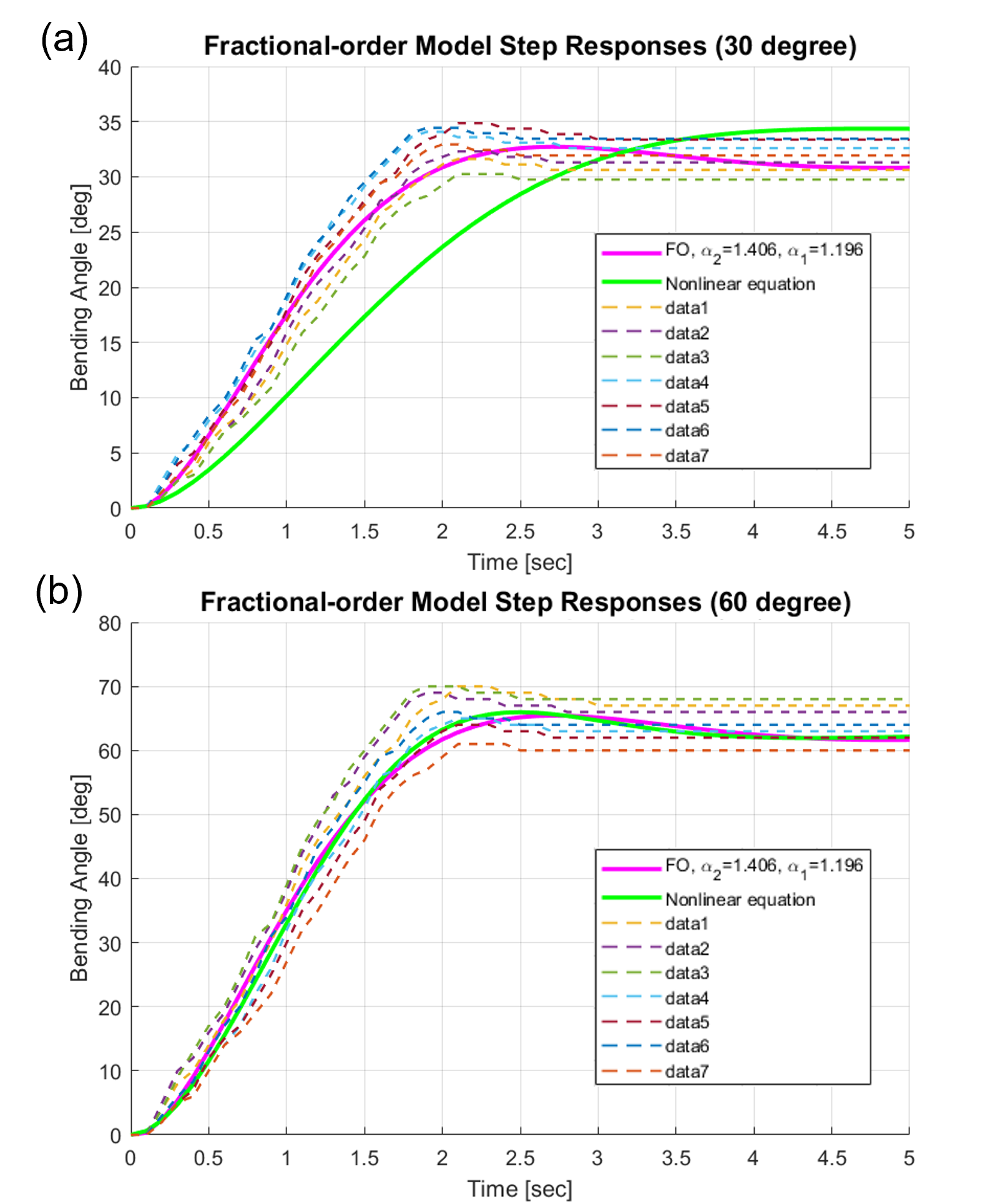}
    \caption{Comparisons between seven experimental data, nonlinear dynamical equation, and fractional-order equation for the soft actuator made of Dragon Skin 20.}
    \label{fig: 4}
    \vspace{-0.15in}
\end{figure}

The system parameters are listed in Table~\ref{tab:Table I} (Dragon Skin 20), and the fractional-order transfer function becomes

\begin{align}
    \begin{split}
    G_{D1}(s) = \frac{0.934}{s^{1.406}+0.794s^{1.196}+1.638}
    \label{eqn: 20}
    \end{split}
\end{align}
The step responses are shown in Fig.~\ref{fig: 4}. In Fig.~\ref{fig: 4}(a), the setpoint is 30 deg. Seven experimental responses are shown and compared with the model prediction to evaluate its accuracy. In addition, the reference of the nonlinear second-order dynamic model (\ref{eqn: 5}) is included for comparison. The root-mean-square (RMS) error between the fractional-order model (\ref{eqn: 20}) and the averaged experimental responses is 1.24 deg (4.13\%), while that of the nonlinear model is 4.75 deg (15.83\%). These results indicate that fractional-order model outperform the nonlinear model for the 30 deg step response.

A second step response test is performed with a larger setpoint of 60 deg to further evaluate model accuracy at higher bending angles (Fig.~\ref{fig: 4}(b)). The RMS error between the nonlinear model and the averaged experimental responses is 3.42 deg (5.7\%), whereas the fractional-order model yields an RMS error of 2.52 deg (4.2\%). Overall, the fractional-order model demonstrates better consistency across both small and large step responses.

\subsubsection{Design 2 Test}
\begin{figure}[t]
    \centering
    \includegraphics[width=205pt]{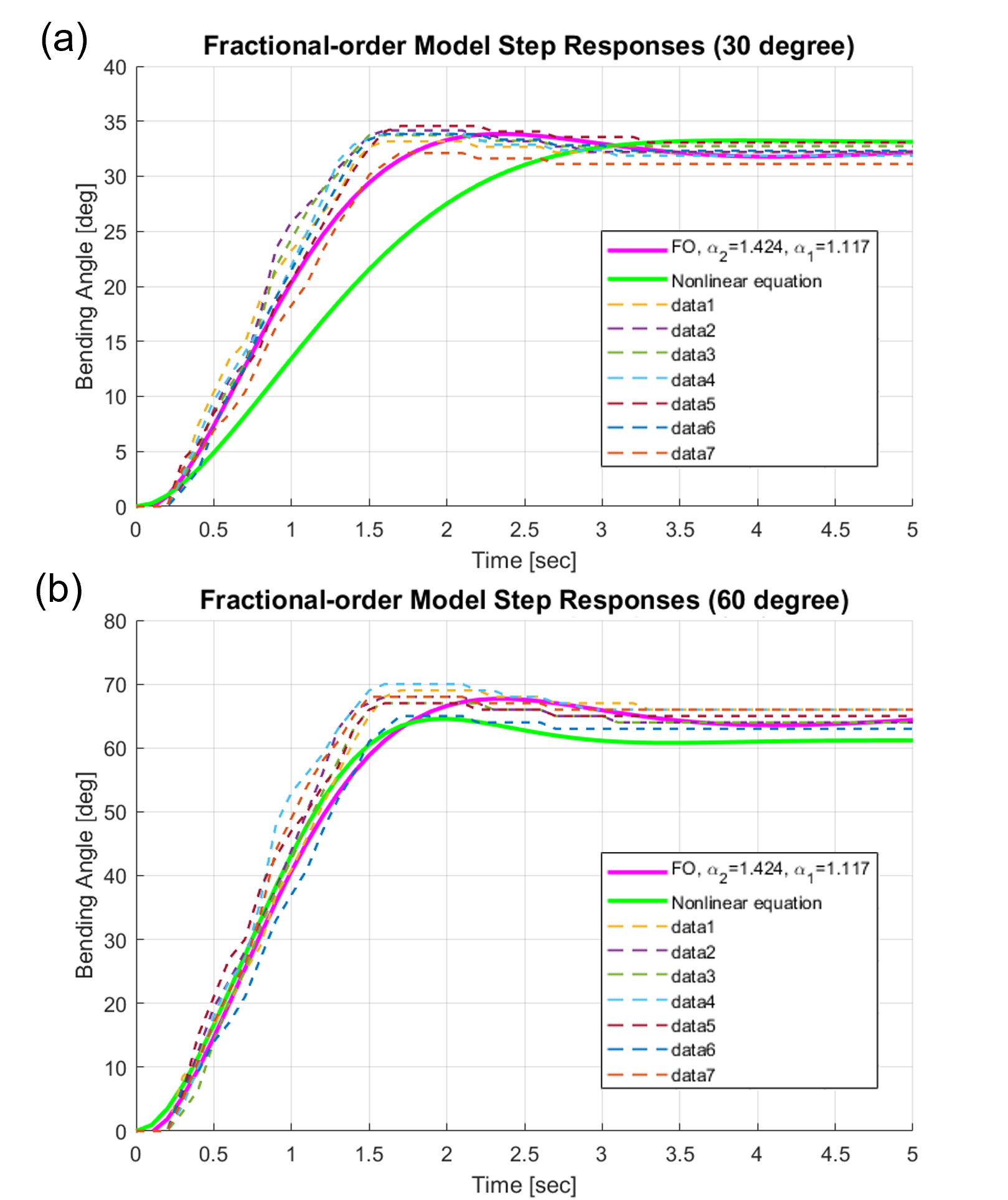}
    \caption{Comparisons between seven experimental data, nonlinear dynamical equation, and fractional-order equation of the soft actuator made of Dragon Skin FX-Pro.}
    \label{fig: 5}
    \vspace{-0.15in}
\end{figure}

The system parameters are listed in Table~\ref{tab:Table I} (Dragon Skin FX-Pro), and the system transfer function is

\begin{align}
    \begin{split}
    G_{D2}(s) = \frac{0.728}{s^{1.424}+0.986s^{1.17}+2.103}
    \label{eqn: 21}
    \end{split}
\end{align}
The step response results are presented in Fig.~\ref{fig: 5}. As shown in Fig.~\ref{fig: 5}(a), a 30 deg setpoint is applied, and seven experimental trials are plotted together with the model prediction for validation. Similar to Design 1 Test, the results obtained from a nonlinear second-order dynamic model (\ref{eqn: 5}) are also included. The fractional-order model (\ref{eqn: 21}) achieves a RMS error of 0.95 deg (3.16 \%) with respect to the averaged experimental data, while the nonlinear model yields an RMS error of 2.79 deg (9.3 \%). 

To further assess the model performance at larger deflections, a second step response test is conducted at a 60 deg setpoint (Fig.~\ref{fig: 5}(b)). In this case, the nonlinear model produces an RMS error of 1.65 deg (2.8 \%), whereas the fractional-order model results in an RMS error of 1.96 deg (3.3 \%). 

To summarize, the fractional-order models capture the dynamic behaviors of both prototypes more accurately than the nonlinear models. The RMS errors remain approximately 3-4 \% for both small and large bending angles under step response tests.

\section{Discussion and Conclusion}\label{sec5}
\subsection{Discussion}\label{subsec51}
\subsubsection{Physical Meaning}\label{subsec511}
Unlike integer-order differential equations, fractional-order models can capture the dynamics of soft materials, whose behavior lies between that of purely elastic solids and purely fluidic media~\cite{hert2020mechanics}. Soft materials often respond at rates that deviate from classical exponential laws, making fractional derivatives a more suitable representation. In this study, $\alpha_1$ and $\alpha_2$ in (\ref{eqn: 15}) are determined via particle swarm optimization. We could interpret that $\tau = Id^{\alpha_2}\theta/dt^{\alpha_2}$ and the damping term is proportional to $a_1'd^{\alpha_1}\theta/dt^{\alpha_1}$. The effectiveness of the proposed modeling approach is validated by two soft actuator prototypes. 

\subsubsection{Limitations of PSO}\label{subsection513}
The limitations of PSO are examined through parameter searches on two prototypes, showing that reliable parameter identification requires at least 4 datasets (i.e., 4 step responses) and a minimum of 200 data points. Using fewer datasets leads to parameter drift and noticeable modeling errors. Although incorporating more data points improves robustness, the parameter estimates remain sensitive to outliers.

\subsection{Conclusion}\label{subsec52}
This study presents a novel fractional-order modeling approach for soft pneumatic actuators. Unlike conventional integer-order models, the proposed method employs fractional-order differential equations to more accurately describe soft actuators' behavior. Key system parameters such as the fractional order, damping ratio, and natural frequency are identified using the particle swarm optimization algorithm. The proposed modeling approach requires less dependence on extensive experimental data compared to previous works. To validate the method, two prototypes made from different soft materials are tested. The root mean square (RMS) errors are around 3-4 \%. Compared with a recently proposed nonlinear model, the fractional-order model improved above 5 \% accuracy for both small and large bending angles during step response experiments.

%


\balance

\bibliographystyle{IEEEtran}
\bibliography{IEEEabrv}


\end{document}